\documentclass[conference]{IEEEtran}
\IEEEoverridecommandlockouts

\usepackage{cite}
\usepackage{amsmath,amssymb,amsfonts}
\usepackage{newtxtext,newtxmath}
\usepackage{graphicx}
\usepackage{textcomp}
\usepackage{xcolor}
\usepackage{booktabs}
\usepackage{url}
\usepackage[hidelinks]{hyperref}
\definecolor{MethodBlue}{RGB}{72,122,158}
\usepackage{cleveref}
\Crefname{definition}{Definition}{Definitions}
\Crefname{proposition}{Proposition}{Propositions}
\Crefname{theorem}{Theorem}{Theorems}
\Crefname{figure}{Fig.}{Figs.}
\Crefname{equation}{Eq.}{Eqs.}
\Crefname{section}{Section}{Sections}
\Crefname{subsection}{Section}{Sections}
\Crefname{subsubsection}{Section}{Sections}
\Crefname{appendix}{Appendix}{Appendices}
\Crefname{algorithm}{Algorithm}{Algorithms}

\newcommand{\Sph}{\mathbb{S}}

\newcommand{\state}{\mathbf{x}}
\newcommand{\control}{\mathbf{u}}
\newcommand{\configuration}{\mathbf{q}}
\newcommand{\velocity}{\mathbf{v}}
\newcommand{\acceleration}{\dot{\mathbf{v}}}
\newcommand{\bodyvelocity}{\mathbf{v}_{B}}
\newcommand{\waterflow}{\mathbf{v}_{w}}
\newcommand{\relativeflow}{\mathbf{r}}
\newcommand{\vehicleflow}{\mathbf{v}_{\mathrm{rel}}}
\newcommand{\flowspeed}{s}
\newcommand{\massmatrix}{\mathbf{M}_{\mathrm{RB}}}
\newcommand{\biasforces}{\mathbf{h}_{\mathrm{RB}}}
\newcommand{\externalforces}{\boldsymbol{\tau}_{\mathrm{ext}}}
\newcommand{\flowdirection}{\mathbf{n}}

\newcommand{\hydrowrench}{\boldsymbol{\tau}_{h}^{C}}
\newcommand{\actuationmatrix}{\mathbf{B}_{u}}
\newcommand{\discretedynamics}{\mathbf{f}_{d}}
\newcommand{\statejacobian}{\mathbf{F}_{x}}
\newcommand{\controljacobian}{\mathbf{F}_{u}}

\newif\ifanonymous
\anonymousfalse
\newcommand*{\videourl}{https://youtu.be/oX3EtKuudBg}
\newcommand*{\videofigure}[1]{\ifanonymous#1\else\href{\videourl}{#1}\fi}
\DeclareRobustCommand{\videocaption}{\ifanonymous\else\space Video: \expandafter\url\expandafter{\videourl}.\fi}
\DeclareRobustCommand{\publicauthornote}{%
  Hefan Wang and Carlos Mastalli are with the Robot Motor Intelligence (RoMI) Lab while Yvan Petillot is part of the Ocean Lab, Heriot-Watt University, Edinburgh, UK.%
  \ifx\videourl\empty\else
    \space Video: \expandafter\url\expandafter{\videourl}.%
  \fi}

\makeatletter
\renewcommand{\@maketitle}{%
  \newpage
  \begingroup
    \par\vskip\IEEEtitletopspace\vskip\IEEEtitletopspaceextra\centering
    \vskip0.2em{\Huge\@title\par}%
    \ifanonymous\else
      \vskip1.7\baselineskip\nointerlineskip
      {\@IEEEauthordefaulttextstyle\@author\par}%
    \fi
  \endgroup}
\makeatother

\begin{document}

\title{Differentiable Hydrodynamics for Whole-Body MPC of Underwater Robots}

\ifanonymous
\author{}
\else
\author{Hefan Wang\quad Yvan Petillot\quad Carlos Mastalli%
\thanks{\publicauthornote}}
\fi

\maketitle
\setlength{\abovedisplayskip}{4pt plus 2pt minus 1pt}
\setlength{\belowdisplayskip}{4pt plus 2pt minus 1pt}
\setlength{\abovedisplayshortskip}{0pt plus 1pt}
\setlength{\belowdisplayshortskip}{3pt plus 1pt minus 1pt}

\begin{abstract}
Underwater vehicle--manipulator systems distribute actuation across thrusters and joint motors. Whole-body model predictive control (MPC), rather than separate vehicle and arm controllers, can coordinate both to extend the workspace and exploit the system's full kinematic capability. Doing so, however, requires a hydrodynamic model and derivatives throughout the prediction horizon. Computational fluid dynamics (CFD) captures geometry-dependent forces and moments for such a model, but is too expensive to evaluate online. We present a differentiable analytical hydrodynamic model combining smooth spherical-harmonic force and moment fields with a coupled added-inertia matrix identified from CFD. It compactly represents velocity and acceleration-dependent loads and provides their derivatives in closed form. These derivatives are propagated through free-floating vehicle--manipulator dynamics for whole-body trajectory optimization and receding-horizon control, jointly optimizing thruster forces and joint torques for reaching and continuous tracking. Held-out CFD tests validate the principal translational-force and rotational-moment fields and the added-inertia response. The analytical Jacobians evaluate faster and avoid the step-size sensitivity of central finite differences. Closed-loop Stonefish experiments demonstrate whole-body MPC in an independent marine simulator. By making geometry-dependent hydrodynamics directly available to gradient-based control, the model incorporates water forces and their sensitivities into motion allocation between the vehicle and arm, rather than treating them only as disturbances.
\end{abstract}

\section{Introduction}

Underwater vehicle--manipulator systems (UVMS) combine vehicle thrusters and manipulator actuators within a strongly coupled floating-base system. Manipulator motion induces vehicle reactions, while vehicle motion reshapes the reachable workspace and end-effector dynamics, with these interactions becoming particularly pronounced in lightweight platforms with comparable vehicle and manipulator inertial scales~\cite{kolano2024,trekel2025}. Conventional decoupled control architectures regulate the vehicle and manipulator separately~\cite{barbalata2018,ding2021}. This is effective when base motion must be suppressed but limits their coordinated use to extend reach, redistribute control effort, or anticipate subsequent maneuvers.

Whole-body model predictive control (MPC) addresses this limitation by optimizing thruster forces and joint torques over the same horizon, following the joint treatment of floating-base and joint dynamics demonstrated in legged robots~\cite{mastalli2022agile,mastalli2023inverse}. The vehicle can move to enlarge the reachable workspace, while the arm can be positioned consistently with vehicle momentum and actuator limits. Marine MPC has been applied to trajectory tracking and predictive rejection of wave disturbances~\cite{molero2011,fernandez2017}. For gradient-based whole-body MPC, every prediction node requires the dynamics and their derivatives, including the water forces and moments that govern acceleration, braking, and turning.

Computational fluid dynamics (CFD) can resolve the geometry-dependent hydrodynamic forces and moments of open-frame vehicles across different directions of translational and rotational motion~\cite{ramirez2016}. Its computational cost, however, precludes direct evaluation within an MPC loop. Conventional linear or diagonal quadratic damping models provide an efficient alternative~\cite{fossen2021}. Diagonal models distinguish resistance along the principal axes but restrict angular variation and omit cross-component coupling. A richer directional representation lets the whole-body optimizer predict how changes in vehicle motion alter the hydrodynamic wrench during planning.

We bridge this gap with a compact model identified from steady and oscillatory CFD. Spherical harmonics represent velocity-dependent forces and moments, while a symmetric added-inertia matrix describes the acceleration of the surrounding water. The model provides closed-form derivatives through the free-floating vehicle--manipulator dynamics. The resulting action model allows the optimizer to decide how vehicle motion, arm motion, directional resistance, and fluid inertia should be combined in a whole-body maneuver.

\begin{figure*}[t]
\centering
\videofigure{\includegraphics[width=0.98\textwidth]{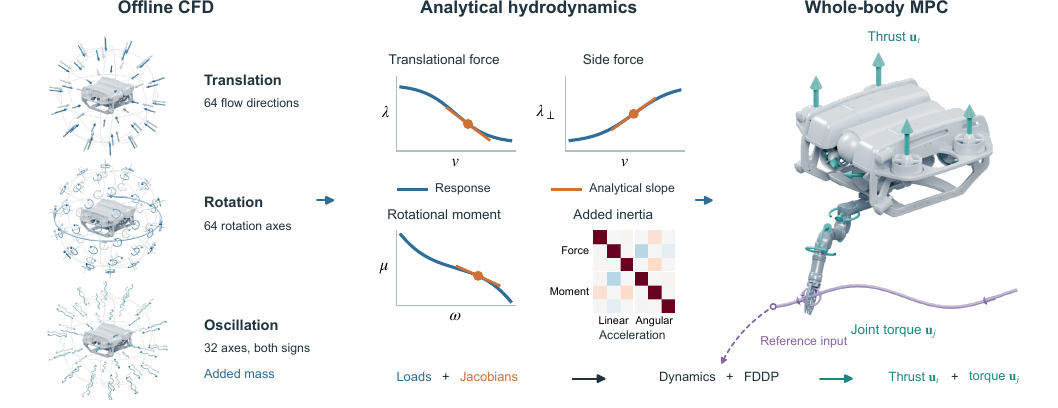}}
\caption{From CFD to whole-body MPC. Offline CFD samples forces and moments during steady translation/rotation and isolates acceleration loads through oscillation. Analytical hydrodynamics fits spherical fields and an inertia matrix with closed-form derivatives. Whole-body MPC predicts multibody motion from these loads; FDDP uses their Jacobians to coordinate thrusters and joints along the tool reference, applying the first control before replanning. Curves and actuation are schematic; matrix entries are $A_{ab}/\sqrt{A_{aa}A_{bb}}$ for the six components.\videocaption}
\label{fig:overview}
\end{figure*}

This work introduces a differentiable hydrodynamics approach for whole-body MPC of underwater robots (see~\Cref{fig:overview}). It relies on three contributions: (i) a compact model of CFD forces and moments with positive directional resistance and positive-definite coupled added inertia; (ii) closed-form hydrodynamic derivatives propagated through free-floating multibody dynamics; and (iii) whole-body trajectory-optimization and MPC experiments in which thrusters and manipulator motors are optimized together, including evaluation in an independently implemented Stonefish plant.

\section{Related Work}

\subsection{Whole-Body Control of UVMSs}

Hierarchical and task-priority controllers coordinate vehicle and manipulator objectives for intervention, landing, and grasping \cite{simetti2015,simetti2018}. Recent lightweight systems and physical emulators further demonstrate that vehicle--arm coupling matters in dynamic tracking and autonomous intervention \cite{palmer2024,rosette2024,trekel2025,bauschmann2025}. MPC has also been used for UVMS interaction and predictive station keeping under disturbances \cite{nikou2018,walker2021}. Our focus is complementary: the vehicle is not treated as a base that must remain fixed. Its motion and the manipulator motion are optimized together, and geometry-dependent hydrodynamics are differentiated inside the prediction model.

\subsection{Hydrodynamic Models for Optimization}

Standard marine models separate rigid-body inertia, added mass, hydrostatics, damping, and actuation \cite{fossen2021}. Virtual captive CFD tests identify direction-dependent ROV forces and moments \cite{ramirez2016}, but the resulting CFD solver cannot be queried during control. Ferrari \emph{et al.} used spherical harmonics to regress ship forces over surge, sway, and yaw-rate directions \cite{ferrari2022,ferrari2025}. We use a spherical representation for the three-dimensional flow seen by an underwater robot, retain both force and moment, impose positive resistance, and derive the Jacobian needed by optimal control. Differentiable underwater simulation also supports system identification and optimal control~\cite{du2021}.

\subsection{Differentiable Multibody Optimal Control}

Recursive rigid-body algorithms and efficient numerical optimization have enabled whole-body MPC for agile legged maneuvers, inverse-dynamics control, and fast constrained motion planning~\cite{mastalli2022agile,mastalli2023inverse,khazoom2024}. Recursive algorithms provide efficient multibody dynamics \cite{featherstone2008}, and \textsc{Pinocchio} supplies their analytical derivatives on general kinematic trees \cite{carpentier2018,carpentier2019}. \textsc{Crocoddyl} combines manifold states, differentiable action models, and shooting solvers for robot optimal control \cite{mastalli2020}. We retain this organization and include buoyancy, directional water loads, and added inertia.

Below, we describe our differentiable hydrodynamics model built from spherical harmonics.

\section{Differentiable Directional Hydrodynamics}
\label{sec:drag}

This section develops a differentiable hydrodynamic model that captures direction-dependent forces, moments, and coupled added inertia from offline CFD data. Smooth spherical representations provide closed-form derivatives for gradient-based trajectory optimization and model predictive control.

\subsection{Model Structure}

\begin{table}[t]
\caption{Directional CFD Sampling}
\label{tab:cfd_sampling}
\centering
\footnotesize
\setlength{\tabcolsep}{3.2pt}
\begin{tabular*}{\columnwidth}{@{\extracolsep{\fill}}lrc@{}}
\toprule
\textbf{Motion} & \textbf{Sampling} & \textbf{Prescribed magnitude}\\
\midrule
Steady translation & 64 directions & Speed: $1\,\mathrm{m\,s^{-1}}$\\
Steady rotation & 64 directions & Speed: $0.5\,\mathrm{rad\,s^{-1}}$\\
\midrule
Oscillatory translation & 32 axes & Peak: $0.25\,\mathrm{m\,s^{-2}}$\\
Oscillatory rotation & 32 axes & Peak: $0.5\,\mathrm{rad\,s^{-2}}$\\
\bottomrule
\end{tabular*}
\par\smallskip\footnotesize Oscillatory magnitudes are peak accelerations. Each axis is traversed in both signs at $0.75$ Hz: four periods, final three retained.
\end{table}

Steady CFD identifies hull force and moment fields; oscillatory CFD identifies the water's added inertia. A coupled symmetric $6\times6$ matrix retains cross-responses with 21 independent parameters; a diagonal matrix uses six. Both enter vehicle--arm dynamics.

\subsection{Relative Twist and CFD Samples}

Hull-fixed axes are $x$ forward, $y$ left, and $z$ up. The hull COM $C$ is the CFD moment reference and rotation center. In these axes, $\bodyvelocity$ and $\waterflow$ are vehicle and water velocities at $C$, and $\boldsymbol\omega$ is angular velocity. Water velocity relative to the vehicle is
\begin{equation}
 \relativeflow=\waterflow-\bodyvelocity,\qquad
 \flowspeed=\|\relativeflow\|,\qquad
 \flowdirection=\relativeflow/\flowspeed\in\Sph^2.
 \label{eq:relative_flow}
\end{equation}
Thus, $\mathbf n$ follows the CFD inlet flow, opposite to vehicle motion through water. For uniform current, the rotational input is
\begin{equation}
 s_\omega=\|\boldsymbol{\omega}\|,\qquad
 \mathbf{n}_\omega=\boldsymbol{\omega}/s_\omega\in\Sph^2.
 \label{eq:relative_rotation}
\end{equation}
Steady cases use 32 quasi-uniform antipodal pairs; oscillations traverse each axis in both signs (Table~\ref{tab:cfd_sampling}). Section~\ref{sec:evaluation} details CFD settings and validation.

At each of $N_s=64$ steady directions, sample $i$ gives force $\boldsymbol\lambda_i$ [N] and moment $\boldsymbol\mu_i$ [N\,m] at speed $s_i$ and direction $\mathbf n_i$. The projector $\mathbf P_i=\mathbf I-\mathbf n_i\mathbf n_i^\top$, with $\mathbf I$ the $3\times3$ identity matrix, separates axial and transverse force coefficients:
\begin{align}
 d_i&=\frac{2\mathbf{n}_i^\top\boldsymbol{\lambda}_i}{\rho s_i^2}>0,
 &\mathbf{t}_i&=\frac{2\mathbf{P}_i\boldsymbol{\lambda}_i}{\rho s_i^2},\label{eq:force_samples}\\
 \mathbf{m}_i&=\frac{2\boldsymbol{\mu}_i}{\rho s_i^2}.
 \label{eq:moment_samples}
\end{align}
Here, $d=C_dA$ is positive drag area, $\mathbf t_i$ the transverse-force coefficient, and $\mathbf m_i$ the moment coefficient. We fit the CFD-identified product $C_dA$, combining drag coefficient and projected area in one directional field.

\subsection{Scalar and Vector Spherical Models}

Like Fourier modes on a circle, spherical harmonics interpolate directions from broad to fine angular structure. Used in illumination and precomputed transport~\cite{ramamoorthi2001,sloan2002}, they give smooth values and derivatives from the same coefficients.

Index $j$ labels the $N_\phi=(L+1)^2$ real basis functions through degree $L$~\cite{atkinson2012}: 25 functions at degree four, versus 64 samples. Fitting log drag area linearly and exponentiating gives positive resistance:
\begin{equation}
 d(\mathbf{n})=\exp\!\left(\sum_{j=1}^{N_\phi}w_j\phi_j(\mathbf{n})\right).
 \label{eq:positive_axial_field}
\end{equation}
The sum represents log drag area; the outer exponential maps this fitted field back to positive drag area. The sample row $\boldsymbol\Phi_i=[\phi_1(\mathbf n_i),\ldots,\phi_{N_\phi}(\mathbf n_i)]$ evaluates the basis at direction $i$. With coefficients $\mathbf w=[w_1,\ldots,w_{N_\phi}]^\top$, $\boldsymbol\Phi_i\mathbf w$ predicts $\log d_i$ via linear ridge fitting:
\begin{equation}
 \begin{aligned}
 &\sum_{i=1}^{N_s}\omega_i\left(\boldsymbol{\Phi}_i\mathbf{w}-\log d_i\right)^2
 +\lambda\|\mathbf{w}\|_2^2,\\
 &\omega_i\propto
 \frac{1}{\max(\sigma_i/d_i,0.005)^2}.
 \end{aligned}
 \label{eq:fit}
\end{equation}
Here, $\omega_i$ is a sample weight, $\sigma_i$ its CFD tail standard deviation, and $\lambda$ the ridge strength. Weights have median one. Grouped cross-validation selects degree four and regularization while keeping antipodal pairs together.

For lift and side force, $\mathbf P=\mathbf I-\mathbf n\mathbf n^\top$ projects onto the tangent plane. Nonconstant modes ($j\geq2$) give $\mathbf g_j=\mathbf P\nabla\phi_j$ for the expansion
\begin{equation}
 \mathbf{t}(\mathbf{n})=
 \sum_{j=2}^{N_\phi}\left[
 a_j\,\mathbf{g}_j(\mathbf{n})+
 b_j\,\mathbf{n}\times\mathbf{g}_j(\mathbf{n})\right]\in\mathbb{R}^3.
 \label{eq:tangent_field}
\end{equation}
Coefficients $a_j,b_j\in\mathbb R$ weight the surface-gradient pattern $\mathbf g_j$ and its $90^\circ$ tangent-plane rotation $\mathbf n\times\mathbf g_j$. Fitted jointly to the CFD transverse-force samples, they remain fixed during control and give $\mathbf t\perp\mathbf n$. The moment is fitted componentwise:
\begin{equation}
 \mathbf{m}(\mathbf{n})=
 \sum_{j=1}^{N_\phi}\mathbf{c}_j\phi_j(\mathbf{n})\in\mathbb{R}^3.
 \label{eq:moment_field}
\end{equation}
Each $\mathbf c_j$ has three moment coefficients about $C$. Both fields use independent ridge fits:
\begin{equation}
 \min_{\boldsymbol{\theta}_k}
 \sum_{i=1}^{N_s}\omega_i\left\|\mathbf{A}_{k,i}\boldsymbol{\theta}_k-\mathbf{y}_{k,i}\right\|_2^2
 +\lambda_k\|\boldsymbol{\theta}_k\|_2^2,
 \qquad k\in\{t,m\}.
 \label{eq:vector_fit}
\end{equation}
Here, $\boldsymbol\theta_k$ stacks the field coefficients; $\mathbf A_{k,i}$ evaluates their scalar or tangent basis at sample $i$, with targets $\mathbf y_{t,i}=\mathbf t_i$ and $\mathbf y_{m,i}=\mathbf m_i$. Each field uses grouped selection. Degree four gives 25 axial, 48 transverse, and 75 moment coefficients.

The translation-induced force and moment are
\begin{equation}
 \boldsymbol{\lambda}_t=\tfrac{1}{2}\rho s^2
 \left[d(\mathbf{n})\mathbf{n}+\mathbf{t}(\mathbf{n})\right],
 \qquad
 \boldsymbol{\mu}_t=\tfrac{1}{2}\rho s^2\mathbf{m}(\mathbf{n}).
 \label{eq:translation_wrench}
\end{equation}
Subscript $t$ denotes translation: $\boldsymbol\lambda_t$ includes axial and transverse force; $\mathbf t$ is only the transverse coefficient. Vehicle velocity relative to water is $\vehicleflow=-\relativeflow=-\flowspeed\flowdirection$, giving
\begin{equation}
 \vehicleflow^\top\boldsymbol{\lambda}_t
 =-\tfrac{1}{2}\rho \flowspeed^3d(\flowdirection)\leq0.
 \label{eq:dissipation}
\end{equation}
Thus, axial drag dissipates translational energy, while the transverse force redirects motion without contributing translational power.

For rotation, subscript $r$ denotes rotation-induced force $\boldsymbol\lambda_r$ and moment $\boldsymbol\mu_r$. The same bases fit the force coefficient $\mathbf f_r$ componentwise and the moment as axial resistance $d_r$ plus a tangent coefficient $\mathbf t_r$:
\begin{align}
 q_r(s_\omega)&=\tfrac{1}{2}\rho
 s_{\mathrm{ref}}^{\,2-\alpha_r}s_\omega^{\alpha_r},\nonumber\\
 \boldsymbol{\lambda}_r&=q_r\mathbf{f}_r(\mathbf{n}_\omega),\nonumber\\
 \boldsymbol{\mu}_r&=-q_r\left[
 d_r(\mathbf{n}_\omega)\mathbf{n}_\omega+
 \mathbf{t}_r(\mathbf{n}_\omega)\right].
 \label{eq:rotation_wrench}
\end{align}
Here, $d_r$ and $\mathbf{t}_r$ follow Eq.~\eqref{eq:positive_axial_field} and Eq.~\eqref{eq:tangent_field}, respectively. The reference speed is $s_{\mathrm{ref}}=0.5\,\mathrm{rad\,s^{-1}}$ and the fitted exponent is $\alpha_r=1.955$. Since $\mathbf{t}_r\perp\mathbf{n}_\omega$, the rotational power satisfies
\begin{equation}
 \boldsymbol{\omega}^{\top}\boldsymbol{\mu}_r
 =-q_r s_\omega d_r(\mathbf{n}_\omega)\leq0.
 \label{eq:rotational_dissipation}
\end{equation}

\begin{figure*}[t]
\centering
\videofigure{\includegraphics[width=0.98\textwidth]{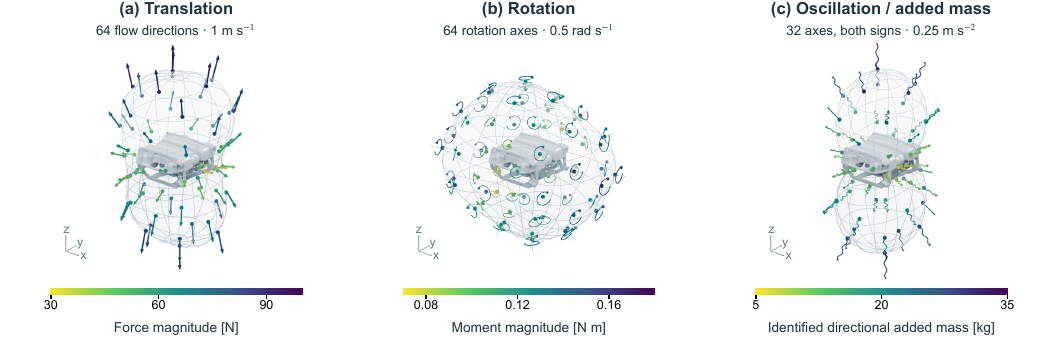}}
\caption{Hull-frame directional responses. Mesh radii scale (a) drag area, (b) rotational resistance, and (c) $\mathbf n^\top\mathbf A_{tt}\mathbf n$ from the constant inertia matrix. Sample glyph color and size encode force [N], moment [N\,m], or identified mass [kg]. Arrows follow force; right-hand arcs follow moment; wave extent denotes mass. Panel (c) shows 32 axes in both signs.\videocaption}
\label{fig:wrench_residuals}
\end{figure*}

The complete quasi-steady body wrench about $C$, expressed in hull-fixed axes, is
\begin{equation}
 \hydrowrench(\relativeflow,\boldsymbol{\omega})=
 \begin{bmatrix}
 \boldsymbol{\lambda}_t(\relativeflow)+\boldsymbol{\lambda}_r(\boldsymbol{\omega})\\
 \boldsymbol{\mu}_t(\relativeflow)+\boldsymbol{\mu}_r(\boldsymbol{\omega})
 \end{bmatrix}.
 \label{eq:hydrodynamic_wrench}
\end{equation}
This separable model preserves the identified pure-motion loads. Eqs.~\eqref{eq:dissipation} and~\eqref{eq:rotational_dissipation} establish pure-motion dissipation; combined motion introduces cross-power $\vehicleflow^\top\boldsymbol\lambda_r+\boldsymbol\omega^\top\boldsymbol\mu_t$. In Fig.~\ref{fig:wrench_residuals}(a,b), larger radii mean greater directional resistance, while sample glyphs show CFD response magnitudes and directions. Panel (c) visualizes the directional response of the constant inertia matrix identified next.

\subsection{Coupled Added Inertia}

Oscillation separates resistance from inertia because velocity and acceleration have different phases. For each CFD-computed force or moment component $y$, we fit
\begin{equation}
 y(t)=b+c_1 u(t)+c_2u(t)|u(t)|-a_{\mathbf n}\dot u(t),
 \label{eq:oscillatory_identification}
\end{equation}
where $u$ is linear or angular velocity along the imposed axis. The first three terms absorb offset and velocity-dependent loads; $a_{\mathbf n}$ identifies one component of the six-dimensional response to acceleration along that axis. Time-weighted regression accounts for adaptive sampling. Uncertainty is estimated across periods for each imposed axis.

The fitted constant matrix is $\mathbf A_C=\mathbf A_C^\top\succ0$. Its $3\times3$ blocks map linear acceleration to force $\mathbf A_{tt}$ and moment $\mathbf A_{rt}$, and angular acceleration to force $\mathbf A_{tr}$ and moment $\mathbf A_{rr}$, with $\mathbf A_{rt}=\mathbf A_{tr}^\top$. The diagonal approximation~\cite{fossen2021} zeros off-diagonal entries. Both use the same data, cycle-covariance weights, and held-out axes. Linear, cross, and angular blocks have units kg, kg\,m, and kg\,m$^2$.

At a common reference point, $[\mathbf p;\mathbf h]=\mathbf A\boldsymbol\nu$ gives fluid linear and angular momentum for relative twist $\boldsymbol\nu=[\mathbf v_{\mathrm{rel}};\boldsymbol\omega]$, with velocity coupling~\cite{fossen2021}
\begin{equation}
 \mathbf b_A(\boldsymbol\nu)=
 \begin{bmatrix}\boldsymbol\omega\times\mathbf p\\
 \mathbf v_{\mathrm{rel}}\times\mathbf p+\boldsymbol\omega\times\mathbf h\end{bmatrix}.
 \label{eq:added_bias}
\end{equation}
Positive definiteness gives nonnegative fluid kinetic energy $\tfrac12\boldsymbol\nu^\top\mathbf A\boldsymbol\nu$, and $\boldsymbol\nu^\top\mathbf b_A=0$. The constant matrix has zero state derivative, while $\mathbf b_A$ depends on velocity.

\subsection{Closed-Form Derivatives}

Gradient-based control needs the sensitivity of water loads to body twist at every trajectory node. We evaluate each $\phi_j$ as a Cartesian solid-harmonic polynomial in $(n_x,n_y,n_z)$, whose gradient and Hessian follow by exact polynomial differentiation. With $\mathbf{P}=\mathbf{I}-\mathbf{n}\mathbf{n}^\top$, the surface derivatives are
\begin{equation}
 \mathbf{g}_j=\mathbf{P}\nabla\phi_j,\qquad
 \mathbf{H}_j=\mathbf{P}\left[\nabla^2\phi_j-
 (\mathbf{n}^\top\nabla\phi_j)\mathbf{I}\right]\mathbf{P}.
 \label{eq:surface_derivatives}
\end{equation}
For a tangent perturbation $\delta\mathbf{n}$,
\begin{align}
 D\mathbf{g}_j[\delta\mathbf{n}]&=
 \mathbf{H}_j\delta\mathbf{n}
 -\mathbf{n}(\mathbf{g}_j^\top\delta\mathbf{n}),\nonumber\\
 D(\mathbf{n}\!\times\!\mathbf{g}_j)[\delta\mathbf{n}]&=
 \delta\mathbf{n}\!\times\!\mathbf{g}_j+
 \mathbf{n}\!\times\!(\mathbf{H}_j\delta\mathbf{n}).
 \label{eq:vector_field_derivatives}
\end{align}
For $\mathbf a=d\mathbf n+\mathbf t$ and $\mathbf m$, these expressions give directional Jacobians $\mathbf J_a$ and $\mathbf J_m$. Relative flow $\relativeflow=s\mathbf n$ gives $D\mathbf n/D\relativeflow=\mathbf P/s$, hence the velocity Jacobians
\begin{align}
 \frac{\partial\boldsymbol{\lambda}_t}{\partial\relativeflow}&=
 \rho\mathbf{a}\relativeflow^\top+
 \frac{\rho s}{2}\mathbf{J}_a\mathbf{P},\nonumber\\
 \frac{\partial\boldsymbol{\mu}_t}{\partial\relativeflow}&=
 \rho\mathbf{m}\relativeflow^\top+
 \frac{\rho s}{2}\mathbf{J}_m\mathbf{P}.
 \label{eq:wrench_jacobian}
\end{align}
The derivatives with respect to $\mathbf{v}_B$ have the opposite sign. For rotation, $D\mathbf{n}_\omega/D\boldsymbol{\omega}=\mathbf{P}_\omega/s_\omega$ and $Dq_r/Ds_\omega=\alpha_r q_r/s_\omega$ give the same radial-plus-surface structure. The complete velocity Jacobian is
\begin{equation}
 \frac{\partial\hydrowrench}{\partial(\relativeflow,\boldsymbol{\omega})}=
 \begin{bmatrix}
 \partial\boldsymbol{\lambda}_t/\partial\relativeflow &
 \partial\boldsymbol{\lambda}_r/\partial\boldsymbol{\omega}\\
 \partial\boldsymbol{\mu}_t/\partial\relativeflow &
 \partial\boldsymbol{\mu}_r/\partial\boldsymbol{\omega}
 \end{bmatrix}.
 \label{eq:complete_wrench_jacobian}
\end{equation}
The spherical fields are smooth; their speed-scaled loads are continuously differentiable at zero, where the branch and its Jacobian vanish. The inertial-load acceleration Jacobian is $-\mathbf A$. Differentiating Eq.~\eqref{eq:added_bias}, with $\delta[\mathbf p;\mathbf h]=\mathbf A\delta\boldsymbol\nu$, gives its velocity Jacobian. The body-frame representation of a fixed world current is also differentiated with respect to orientation.

\section{Whole-Body Optimal Control}
\label{sec:wholebody}

Reaching tasks prescribe tool poses with brief low-velocity dwells; tracking specifies a time-varying tool pose. The optimizer distributes motion between vehicle and arm without a prescribed vehicle trajectory.

State $\state=(\configuration,\velocity)$ comprises base pose, joint positions, and generalized velocity; $\control=[\mathbf u_t;\mathbf u_j]$ stacks eight thruster forces [N] and four joint torques [N\,m]. Costs penalize tracking, posture, velocity, actuation, acceleration, and control variation. Base orientation uses a unit quaternion; pose and feedback errors use manifold differences.

\subsection{Hydrostatics and Actuation}

Buoyancy uses total displaced volume $V$ and a center fixed in the vehicle frame. Its world-frame force and body-frame wrench are
\begin{align}
 \boldsymbol{\lambda}_{b}^{W}&=\rho Vg\mathbf{e}_z^W,\nonumber\\
 \boldsymbol{\tau}_{b}^{B}&=
 \begin{bmatrix}
 \mathbf{R}_{BW}\boldsymbol{\lambda}_{b}^{W}\\
 \mathbf{r}_{CB}\times\mathbf{R}_{BW}\boldsymbol{\lambda}_{b}^{W}
 \end{bmatrix},
 \label{eq:buoyancy}
\end{align}
where $\mathbf{r}_{CB}$ locates the center of buoyancy relative to the vehicle reference. The floating-base Jacobian maps this wrench to generalized forces. Gravity includes every link; buoyancy is concentrated at the assembled system's fixed nominal center, approximating configuration-dependent link buoyancy and joint moments. Its orientation derivative is analytical.

Thruster $a$, at base-relative position $\mathbf r_a$ with unit axis $\mathbf d_a$, contributes wrench $[\mathbf d_a;\mathbf r_a\times\mathbf d_a]u_{t,a}$. Matrix $\mathbf B_u$ maps $[\mathbf u_t;\mathbf u_j]$ to generalized forces; robot descriptions supply actuator limits.

\subsection{Discrete Action Model and Derivatives}

Recursive kinematics provides link poses, velocities, and Jacobians. Rigid-body mass $\mathbf M_{RB}$ sums $m_l\mathbf J_{v,l}^\top\mathbf J_{v,l}+\mathbf J_{\omega,l}^\top\mathbf I_l\mathbf J_{\omega,l}$ over links $l$, with rotational inertia $\mathbf I_l$; bias $\mathbf h_{RB}$ projects their Newton--Euler inertial and gravity loads through the Jacobians~\cite{featherstone2008}. We propagate pose and velocity sensitivities through the tree and analytically differentiate these link contributions.

Matrix $\mathbf E$ selects the floating-base twist; $\mathbf H$ transports it from base origin $B$ to COM $C$: $[\mathbf v;\boldsymbol\omega]\mapsto[\mathbf v+\boldsymbol\omega\times\mathbf r_{BC};\boldsymbol\omega]$. Thus $\mathbf A_B=\mathbf H^\top\mathbf A_C\mathbf H$ and $\mathbf M=\mathbf M_{RB}+\mathbf E^\top\mathbf A_B\mathbf E$. The dual transform maps body wrench to generalized forces: $\boldsymbol\tau_h=\mathbf E^\top\mathbf H^\top\hydrowrench$.

Steady CFD already captures pure-motion pressure and viscous loads. To retain them while adding the identified acceleration response and mixed-motion inertial coupling, we subtract the pure-motion terms from $\mathbf b_A$.

For a uniform world current expressed in the body frame as $\waterflow$, the relative base twist is $\boldsymbol\nu=\mathbf E\velocity-[\waterflow;\mathbf0]$. To extract its pure-motion components about the CFD reference point $C$, where $\mathbf v_{\mathrm{rel},C}=-\relativeflow$, we split the twist at $C$ and transport each component back to $B$:
\[
 \boldsymbol\nu_T=\mathbf H^{-1}[\mathbf v_{\mathrm{rel},C};\mathbf0],\qquad
 \boldsymbol\nu_R=\mathbf H^{-1}[\mathbf0;\boldsymbol\omega].
\]
Their sum recovers $\boldsymbol\nu$. The resulting additional bias is
\begin{equation}
 \begin{aligned}
 \widetilde{\mathbf b}_A={}&\mathbf b_A(\boldsymbol\nu)
 -\mathbf b_A(\boldsymbol\nu_T)-\mathbf b_A(\boldsymbol\nu_R)\\
 &+\mathbf A_B[\boldsymbol\omega\times\waterflow;\mathbf0].
 \end{aligned}
 \label{eq:inertia_partition}
\end{equation}
Every $\mathbf b_A$ in Eq.~\eqref{eq:inertia_partition} uses $\mathbf A_B$. In still water and constant pure translation or rotation about $C$, the first three terms cancel, recovering the steady CFD wrench. Under combined motion, they retain the bilinear coupling of the identified inertia. The last term accounts for body-frame current variation: $\dot{\waterflow}=-\boldsymbol\omega\times\waterflow$. This closure therefore combines the identified acceleration response with the steady fits. The resulting acceleration satisfies
\begin{equation}
 \mathbf M\acceleration=
 \actuationmatrix\control+\boldsymbol{\tau}_{b}
 +\boldsymbol{\tau}_{h}-\biasforces-\mathbf E^\top\widetilde{\mathbf b}_A.
 \label{eq:implemented_forward_dynamics}
\end{equation}
Linearizing Eq.~\eqref{eq:implemented_forward_dynamics} yields
\begin{equation}
 \mathbf M\,\delta\acceleration=
 \delta\externalforces
 -\delta\biasforces
 -\mathbf E^\top\delta\widetilde{\mathbf b}_A
 -\delta\massmatrix\,\acceleration,
 \label{eq:dynamics_derivative}
\end{equation}
where $\delta\boldsymbol\tau_{\mathrm{ext}}$ includes actuation, buoyancy, and steady-load variations. The link derivatives supply $\delta\mathbf h_{RB}$ and $\delta\mathbf M_{RB}\acceleration$; hydrodynamic Jacobians supply load and added-bias variations. Matrices $\mathbf A_B,\mathbf E$ are constant in body-tangent coordinates. At fixed state, ideal actuation gives $\mathbf M\delta\dot{\mathbf v}=\mathbf B_u\delta\mathbf u$. Acceleration and all sensitivity columns reuse one factorization without explicit inversion.

Prediction uses semi-implicit Euler integration with step $\Delta t$,
\begin{align}
 \velocity_{k+1}&=\velocity_k+\Delta t\acceleration_k,\\
 \configuration_{k+1}&=\configuration_k\oplus
 (\Delta t\velocity_{k+1}),
 \label{eq:integration}
\end{align}
where $\oplus$ integrates on the configuration manifold. Nodes span 30 ms; Stonefish helix prediction composes three 10-ms steps at constant control. This defines the discrete state-transition map $\state_{k+1}=\discretedynamics(\state_k,\control_k)$, whose Jacobians $\statejacobian=\partial\discretedynamics/\partial\state$ and $\controljacobian=\partial\discretedynamics/\partial\control$ enter the backward pass~\cite{mastalli2022feasibility}.

\subsection{Shooting Solver and Receding-Horizon Use}

Feasibility-driven differential dynamic programming (FDDP)~\cite{mastalli2020,mastalli2022feasibility} solves the multiple-shooting problem: derivatives define a backward-pass search direction, tested by nonlinear line-search rollouts. The action model also supports SQP methods such as OdynSQP~\cite{rojas2026odyn}; our experiments use FDDP.

For MPC, a rollout of shifted controls from the measured state initializes FDDP; the tail holds the last command. Stonefish helix warm rollouts also use shifted nominal states and feedback gains. Gains are refreshed after control projection; the first node is executed before replanning.

An ideal thruster applies the command immediately. We also test an augmented state with actual thruster forces $T_a$:
\begin{equation}
 \dot{T}_a=(u_{t,a}-T_a)/\tau_T,
 \label{eq:thruster_dynamics}
\end{equation}
We prescribe $\tau_T=0.18$ s in prediction and simulation: six 30-ms nodes per time constant and approximately $0.54$ s to reach $95\%$ of a step before rate limiting. This simulation setting introduces finite actuator response~\cite{walker2020}. Interval-averaged actual thrust replaces $u_{t,a}$; thrust/command rates are limited to $45/70\,\mathrm{N\,s^{-1}}$. A quadratic variation cost discourages rapid changes. Projection Jacobians are evaluated within each smooth branch.

\section{Evaluation}
\label{sec:evaluation}

Evaluation links load prediction and derivative accuracy to whole-body motion. The simulated BlueROV2 Heavy--Reach Alpha 5 configuration is given in Table~\ref{tab:platform}. Figs.~\ref{fig:whole_body_tracking}--\ref{fig:steady_allocation} examine offline motion allocation; Fig.~\ref{fig:stonefish} tests measured-state MPC in Stonefish. Videos replay the corresponding saved states.

\begin{table}[t]
\caption{Simulated Platform Parameters}
\label{tab:platform}
\centering\footnotesize
\setlength{\tabcolsep}{2.2pt}
\begin{tabular*}{\columnwidth}{@{\extracolsep{\fill}}lp{0.55\columnwidth}@{}}
\toprule
\textbf{Quantity} & \textbf{Value} \\
\midrule
\multicolumn{2}{@{}l}{\textit{BlueROV2 Heavy: eight T200 thrusters}}\\
Envelope $(x,y,z)$ & $0.4572\times0.5750\times0.2539$ m \\
Mass / displacement & $13.0$ kg / $13.50$ L \\
Principal inertia & $(0.26,0.23,0.37)\,\mathrm{kg\,m^2}$ \\
\midrule
\multicolumn{2}{@{}l}{\textit{Reach Alpha 5: four controlled revolute joints}}\\
Mass / displacement & $1.4124$ kg / $0.46$ L \\
Maximum reach & $0.3263$ m \\
Neutral projected areas & $(0.0201,0.0260,0.0220)\,\mathrm{m^2}$ \\
\bottomrule
\end{tabular*}
\end{table}

\subsection{Predicting Loads at Unseen Directions}
\label{subsec:cfd_eval}

Prediction is tested at directions excluded from fitting. Each 64-direction steady campaign uses eight folds of 56 training and 8 test directions, keeping antipodes together. Candidates share samples, weights, and folds; hyperparameters use training data only. Added inertia keeps translation/rotation pairs together: 28 training and 4 test axes per fold, from 32 axes.

The CFD reference uses \textsc{OpenFOAM}~13, RANS $k$--$\omega$ SST, and water density $998.2\,\mathrm{kg\,m^{-3}}$. Steady SIMPLE cases use 2.55 million cells with inflow or a rotating reference-frame region; oscillatory moving meshes use 0.427 million cells. Steady loads average iterations 401--500 after stationary-bias removal. A medium--fine grid comparison changes force by $13.45\%$ in one tested direction. Meshes, reference coordinates, and grid, cycle, and frequency checks accompany the data.

Table~\ref{tab:cfd} compares our fields with smooth componentwise spherical harmonics (SH, 25 degree-four modes per component)~\cite{atkinson2012} and spherical radial basis functions (RBF, 24 kernels per component)~\cite{wendland2005}. Our axial positivity and orthogonal tangent field enforce pure-motion dissipation, Eqs.~\eqref{eq:dissipation} and~\eqref{eq:rotational_dissipation}. Translation also compares three directionally blended quadratic maps and a no-log axial ablation. NRMSE divides the unweighted pooled held-out residual norm by the CFD target norm.

\begin{table}[t]
\caption{Model Structure and Held-Out CFD Prediction Error}
\label{tab:cfd}
\centering
\footnotesize
\setlength{\tabcolsep}{2.1pt}
\begin{tabular*}{\columnwidth}{@{\extracolsep{\fill}}p{0.34\columnwidth}p{0.28\columnwidth}rr@{}}
\toprule
\textbf{Steady-load model} & \textbf{Built-in structure} & $\boldsymbol\lambda_t$ [\%] & $\boldsymbol\mu_r$ [\%]\\
\midrule
Structured SH (Ours) & Axial $>0$ + tangent & $10.03$ & $10.35$\\
Componentwise SH & Cartesian SH fits & $10.32$ & $9.85$\\
Spherical RBF & Radial kernel fits & $8.74$ & $9.62$\\
\bottomrule
\end{tabular*}
\par\smallskip\footnotesize Additional translation-force comparisons: cross-coupled quadratic, $10.79\%$; no-log ablation, $8.77\%$ NRMSE.
\par\smallskip
\begin{tabular*}{\columnwidth}{@{\extracolsep{\fill}}lrr@{}}
\toprule
\textbf{Acceleration-response NRMSE} & Coupled (Ours) & Diagonal~\cite{fossen2021}\\
\midrule
Linear acceleration $\to$ force & $\mathbf{0.242\%}$ & $1.384\%$\\
Linear acceleration $\to$ moment & $\mathbf{1.524\%}$ & $100\%$\\
Angular acceleration $\to$ force & $\mathbf{2.109\%}$ & $100\%$\\
Angular acceleration $\to$ moment & $\mathbf{0.0817\%}$ & $1.846\%$\\
\bottomrule
\end{tabular*}
\par\smallskip\footnotesize Lower NRMSE is better; bold marks the lower inertia error. $\boldsymbol\lambda_t$: translation-induced force; $\boldsymbol\mu_r$: rotation-induced moment.
\end{table}

\begin{figure}[t]
\centering
\includegraphics[width=0.98\columnwidth]{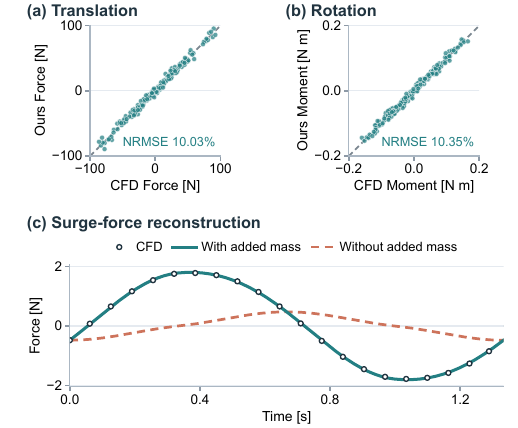}
\caption{CFD validation. (a,b) Ours versus CFD: 192 components per panel from 64 held-out vectors. Dashed lines: exact agreement; percentages: vector NRMSE. (c) Surge reconstruction over the last of three fitted periods, $t=4$--$5.33$ s, displayed from zero. Teal: velocity and acceleration loads; coral: velocity loads.}
\label{fig:held_out_cfd}
\label{fig:added_inertia}
\end{figure}

The held-out predictions in Fig.~\ref{fig:held_out_cfd}(a,b) give a mean force-direction error of $4.08^\circ$ and a median moment-direction error of $3.79^\circ$.

Translation-induced moment has $25.41\%$ NRMSE and $0.72\,\mathrm{N\,m}$ RMSE. Rotation-induced force has $0.0928$ N RMSE ($71.78\%$ NRMSE), compared with $0.1293$ N for a zero-force predictor.

Coupled inertia captures the force--moment cross-responses absent from the diagonal model (Table~\ref{tab:cfd}); all cross-validation fits are positive definite. Acceleration loads recover waveform amplitude and phase in Fig.~\ref{fig:added_inertia}(c), where omitting them gives $97$--$100\%$ NRMSE. Separate leave-one-period-out checks give at most $1.33\%$ principal-response error.

\subsection{Derivatives and Online Computation}
\label{subsec:derivative_eval}

The complete action derivatives agree with central differences at six random states: at step $10^{-5}$, maximum absolute discrepancies are $7.45\times10^{-9}$ for $\mathbf{F}_x$ and $3.66\times10^{-9}$ for $\mathbf{F}_u$. Tests include current, buoyancy, added inertia, manifold integration, Eq.~\eqref{eq:inertia_partition}, reference transformations, and the thruster-state action. Control perturbations stay within a projection branch.

\begin{figure}[t]
\centering
\includegraphics[width=0.98\columnwidth]{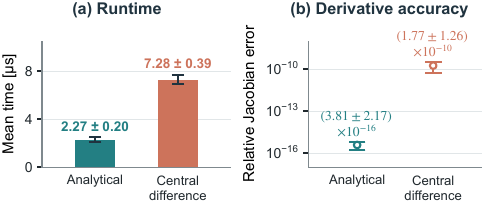}
\caption{Translational hydrodynamic kernel: force, moment, and velocity Jacobian, mean $\pm$ SD. (a) Identical interface: 10,000 warm-up and 10,000 timed calls per method; SD across 100 batch means (i7-14700K). (b) Error against a 90-digit reference over 100 velocities: RMS of force- and moment-Jacobian relative Frobenius errors. Points and error bars use a logarithmic axis.}
\label{fig:action_derivatives}
\end{figure}

Fig.~\ref{fig:action_derivatives} compares the hydrodynamic kernel itself. Analytical derivatives are faster and more accurate over $0.25$--$2$ m/s against a reference checked at 70/90 digits. Central differences use $h_\ell=10^{-6}\max(1,|v_\ell|)$ m/s.

Full action derivatives average $0.1450$ ms with steady loads and $0.1650$ ms with inertia (20 warm-up, 300 timed calls). Table~\ref{tab:timing} then reports FDDP solves and policy validation during MPC. Matched-model tests use 30-ms integration, with and without the thruster dynamics in Eq.~\eqref{eq:thruster_dynamics}.

\begin{table}[t]
\caption{MPC Solve and Policy-Validation Time}
\label{tab:timing}
\centering\footnotesize
\setlength{\tabcolsep}{2pt}
\begin{tabular*}{\columnwidth}{@{\extracolsep{\fill}}lrrrrr@{}}
\toprule
\textbf{Simulation} & $N$ & \textbf{Horizon} & \textbf{Iter. cap} & \textbf{Median} & \textbf{p95}\\
 & & [s] & & [ms] & [ms]\\
\midrule
Matched model & 24 & 0.72 & 2 & 20.19 & 22.89\\
Matched model$^*$ & 8 & 0.24 & 4 & 18.14 & 19.35\\
Matched model & 32 & 0.96 & 4 & 33.69 & 49.39\\
Stonefish reach & 24 & 0.72 & 4 & 50.20 & 79.58\\
Stonefish helix & 24 & 0.72 & 4 & 115.66 & 173.41\\
\bottomrule
\end{tabular*}
\par\smallskip\footnotesize Node/replanning periods: 30 ms simulated. Feedback: every 30 ms (matched) or 15 ms (Stonefish). $^*$Thruster lag. Timing includes FDDP, projected rollout and gain refresh; excludes warm-start preparation, observer, communication and simulation. Samples: 232 per matched test after 8 warm-ups; 400 reaching and 3600 helix updates in Stonefish, including startup. p95: 95th percentile.
\end{table}

\begin{figure*}[t]
\centering
\videofigure{\includegraphics[width=0.98\textwidth]{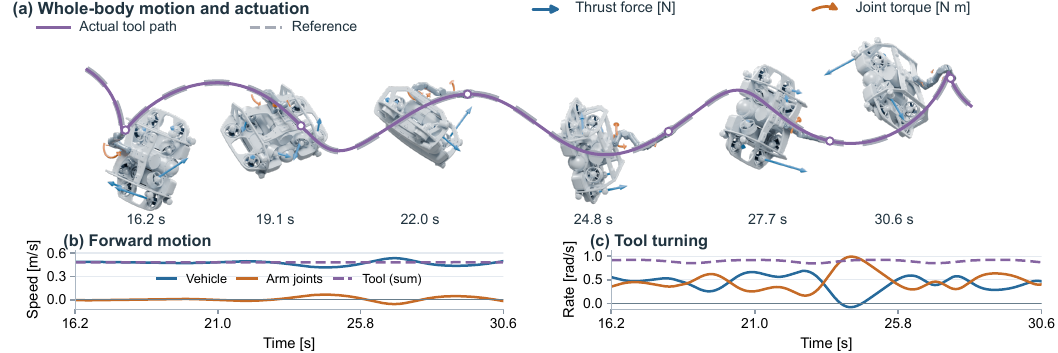}}
\caption{Offline whole-body trajectory optimization with added inertia. (a) Six poses from a 54-s helix; only the tool reference is prescribed. Blue arrows: thrust; orange arcs: joint torque; circles: tool. Signed vehicle/arm contributions sum to tool motion: (b) speed along the reference tangent; (c) angular velocity along the instantaneous tool rotation axis.\videocaption}
\label{fig:whole_body_tracking}
\end{figure*}

\begin{figure}[t]
\centering
\includegraphics[width=0.98\columnwidth]{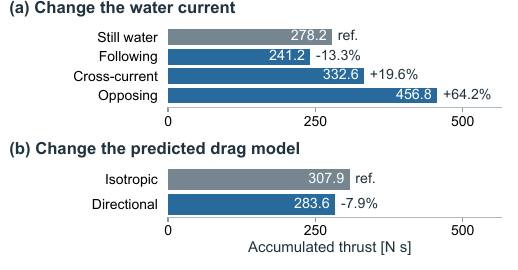}
\caption{Offline absolute thrust impulse [N\,s]. (a) Tracking with varying current, fixed directional model: integrate local $0$--$4.8$ s (source $9$--$13.8$ s); nonzero flow: $0.35\,\mathrm{m\,s^{-1}}$; percentages relative to still water. (b) Still-water reaching with varying drag model: integrate $0$--$9$ s; percentages relative to isotropic drag. Both compare quasi-steady loads with added inertia set to zero.}
\label{fig:steady_allocation}
\end{figure}

\begin{figure*}[t]
\centering
\includegraphics[width=0.98\textwidth]{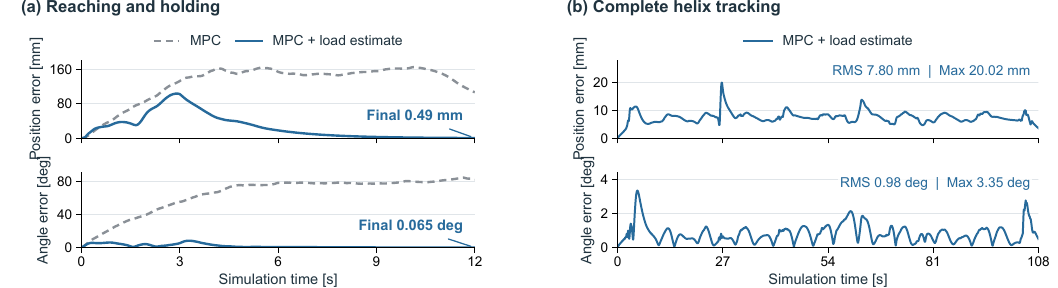}
\caption{Measured-state MPC in Stonefish, with residual forces and torques estimated from measured motion. (a) Side-current reaching ($0.2\,\mathrm{m\,s^{-1}}$): with/without estimation, 400 updates each. (b) Complete still-water helix: estimation and feedback warm rollouts, 3600 updates.}
\label{fig:stonefish}
\end{figure*}

\subsection{Vehicle--Arm Coordination and Thrust Allocation}

With only the tool trajectory prescribed, each offline task optimizes vehicle--arm motion in one segment-initialized shooting problem with a box--sphere self-collision cost. Comparisons fix tasks, costs, limits, and load reference points.

In Fig.~\ref{fig:whole_body_tracking}, the vehicle supplies most forward tool motion ($0.48$ m/s), while turning is shared. Panel (c) projects $\boldsymbol\omega_{\mathrm{tool}}=\mathbf R_{WB}\boldsymbol\omega+\mathbf J_{\omega,\mathrm{arm}}\dot{\mathbf q}_{\mathrm{arm}}$ onto its instantaneous axis, all in world coordinates. Near 24.8 s, the arm turns while the vehicle counter-rotates. Position RMS is $4.48$ mm in the illustrated interval and $4.06$ mm over the full 54-s offline helix, with a $6.16$-mm maximum. Full-trajectory orientation RMS is $0.00874$ rad; peak commands are $31.04$ N and $1.427\,\mathrm{N\,m}$ with ideal thrust.

Fig.~\ref{fig:steady_allocation} examines changes in thrust allocation. Absolute thrust impulse, $I_T=\int_0^{t_f}\sum_{a=1}^{8}|T_a(t)|\,dt$, integrates the eight force magnitudes over each comparison interval, in N\,s. Following current reduces it by $13.3\%$ relative to still water; in the separate rest-to-rest task, the directional model reduces it by $7.9\%$ relative to isotropic drag.

Geometry, three fixed-arm CFD comparisons, and a trajectory-based joint-motion screen assess the small arm's loads. With $C_d=1.2$, the latter estimates $0.148$ N RMS and $2.57$ N maximum, or $0.66\%$ and $6.04\%$ of vehicle-force measures. The control model uses hull hydrodynamics and lumped buoyancy over these tested motions.

\subsection{Transfer to an Independent Marine Simulator}

Measured-state MPC is tested against Stonefish's independent buoyancy, drag, and geometric added inertia. The predictor uses CFD fields and coupled inertia over 24 nodes (0.72 s), with up to four FDDP iterations and neutral controls for the initial solve. Stonefish waits for each optimization, then advances 30 ms with 1-ms integration and 15-ms feedback.

A load-bias observer estimates residual generalized forces and torques from frame-consistent measured-minus-predicted velocity $\delta\velocity$. Prediction uses applied controls and the current estimate. A $0.15$-s-filtered correction $\mathbf M\delta\velocity/\Delta t$ incrementally updates the estimate added to the right-hand side of Eq.~\eqref{eq:implemented_forward_dynamics}. It is fixed over each horizon and bounded by $50$ N, $5$ N\,m (base), and $2$ N\,m (joints). Hydrodynamic coefficients remain fixed.

In Fig.~\ref{fig:stonefish}(a), estimation reduces final error from $106.44$ to $0.49$ mm and from $82.12^\circ$ to $0.065^\circ$. Last-3-s errors stay below $2.92$ mm and $0.272^\circ$.

Fig.~\ref{fig:stonefish}(b) follows all twelve coils in 108 s using the feedback warm rollout and prediction substeps of Section~\ref{sec:wholebody}. All 3600 replans complete without fallback or actuator saturation. Position RMS/maximum are $7.80/20.02$ mm; orientation RMS/maximum are $0.98/3.35^\circ$. The moving-target errors remain bounded throughout the loop.

\section{Discussion}

The model and feedback serve complementary roles: directional loads and coupled inertia shape predicted vehicle--arm motion, while replanning and load-bias estimation correct execution discrepancies. The allocation and Stonefish results thus connect hydrodynamic prediction to whole-body control under simulator--predictor mismatch.

\section{Conclusion}

CFD-identified directional loads and coupled added inertia provide closed-form derivatives for whole-body control. Held-out errors are about $10\%$ for the principal steady-load fields and below $2.2\%$ for all four added-inertia responses. The analytical Jacobians avoid finite-difference step selection and propagate these loads through vehicle--manipulator dynamics. With only tool references prescribed, optimization coordinates vehicle and arm motion and changes thrust allocation with directional resistance. In Stonefish, measured-state MPC with load-bias estimation tracks a complete 108-s helix with $7.80$-mm position RMS error. These results connect geometry-dependent hydrodynamic prediction to coordinated planning and feedback control in an independent simulator.


\begin{thebibliography}{00}
\fontsize{7.9}{8.5}\selectfont

\bibitem{kolano2024}
H. Kolano, E. Palmer, and J. R. Davidson, ``The coupling effect: Experimental validation of the fusion of Fossen and Featherstone to simulate UVMS dynamics in Julia,'' in \emph{Proc. OCEANS}, Halifax, Canada, 2024.

\bibitem{trekel2025}
N. Trekel, N. Bauschmann, T. L. Alff, D. A. Duecker, S. Haddadin, and R. Seifried, ``Dynamic end effector trajectory tracking for small-scale underwater vehicle--manipulator systems: Modeling, control, and experimental validation,'' in \emph{Proc. IEEE Int. Conf. Robot. Autom.}, 2025, pp.~4145--4151.

\bibitem{barbalata2018}
C. Barbalata, M. W. Dunnigan, and Y. Petillot, ``Coupled and decoupled force/motion controllers for an underwater vehicle-manipulator system,'' \emph{J. Mar. Sci. Eng.}, vol.~6, no.~3, Art.~no.~96, 2018.

\bibitem{ding2021}
N. Ding, Y. Tang, Z. Jiang, Y. Bai, and S. Liang, ``Station-keeping control of autonomous and remotely-operated vehicles for free floating manipulation,'' \emph{J. Mar. Sci. Eng.}, vol.~9, no.~11, Art.~no.~1305, 2021.

\bibitem{mastalli2022agile}
C. Mastalli, W. Merkt, G. Xin, J. Shim, M. Mistry, I. Havoutis, and S. Vijayakumar, ``Agile maneuvers in legged robots: A predictive control approach,'' arXiv:2203.07554, 2022.

\bibitem{mastalli2023inverse}
C. Mastalli, S. P. Chhatoi, T. Corb\`eres, S. Tonneau, and S. Vijayakumar, ``Inverse-dynamics MPC via nullspace resolution,'' \emph{IEEE Trans. Robot.}, vol.~39, no.~4, pp.~3222--3241, 2023.

\bibitem{molero2011}
A. Molero, R. Dunia, J. Cappelletto, and G. Fernandez, ``Model predictive control of remotely operated underwater vehicles,'' in \emph{Proc. IEEE Conf. Decision Control Eur. Control Conf.}, 2011, pp.~2058--2063.

\bibitem{fernandez2017}
D. C. Fern\'andez and G. A. Hollinger, ``Model predictive control for underwater robots in ocean waves,'' \emph{IEEE Robot. Autom. Lett.}, vol.~2, no.~1, pp.~88--95, 2017.

\bibitem{ramirez2016}
J. A. Ram\'irez-Mac\'ias, P. Brongers, S. R\'ua, and R. E. V\'asquez, ``Hydrodynamic modelling for the remotely operated vehicle Visor3 using CFD,'' \emph{IFAC-PapersOnLine}, vol.~49, no.~23, pp.~187--192, 2016.

\bibitem{fossen2021}
T. I. Fossen, \emph{Handbook of Marine Craft Hydrodynamics and Motion Control}, 2nd~ed. Hoboken, NJ, USA: Wiley, 2021.

\bibitem{simetti2015}
E. Simetti and G. Casalino, ``Whole body control of a dual arm underwater vehicle manipulator system,'' \emph{Annu. Rev. Control}, vol.~40, pp.~191--200, 2015.

\bibitem{simetti2018}
E. Simetti, G. Casalino, F. Wanderlingh, and M. Aicardi, ``Task priority control of underwater intervention systems: Theory and applications,'' \emph{Ocean Eng.}, vol.~164, pp.~40--54, 2018.

\bibitem{palmer2024}
E. Palmer, C. Holm, and G. A. Hollinger, ``Angler: An autonomy framework for intervention tasks with lightweight underwater vehicle manipulator systems,'' in \emph{Proc. IEEE Int. Conf. Robot. Autom.}, 2024, pp.~6126--6132.

\bibitem{rosette2024}
M. Rosette, H. Kolano, C. Holm, G. A. Hollinger, A. Marburg, M. Pickett, and J. R. Davidson, ``WAVE: An open-source underWater arm--vehicle emulator,'' in \emph{Proc. IEEE Int. Conf. Robot. Autom.}, 2024, pp.~1505--1511.

\bibitem{bauschmann2025}
N. Bauschmann, V. Lenz, R. Seifried, and D. A. Duecker, ``Experimental open-source framework for underwater pick-and-place studies with lightweight UVMS---an extensive quantitative analysis,'' in \emph{Proc. IEEE/RSJ Int. Conf. Intell. Robots Syst.}, 2025, pp.~5307--5314.

\bibitem{nikou2018}
A. Nikou, C. K. Verginis, and D. V. Dimarogonas, ``A tube-based MPC scheme for interaction control of underwater vehicle manipulator systems,'' in \emph{Proc. IEEE/OES Auton. Underwater Veh. Symp.}, 2018, pp.~1--6.

\bibitem{walker2021}
K. L. Walker, R. Gabl, S. Aracri, Y. Cao, A. A. Stokes, A. Kiprakis, and F. Giorgio-Serchi, ``Experimental validation of wave induced disturbances for predictive station keeping of a remotely operated vehicle,'' \emph{IEEE Robot. Autom. Lett.}, vol.~6, no.~3, pp.~5421--5428, 2021.

\bibitem{ferrari2022}
V. Ferrari, S. Sutulo, and C. Guedes Soares, ``Spherical harmonic expansion of hydrodynamic hull forces,'' in \emph{Trends in Maritime Technology and Engineering}, C. Guedes Soares and T. A. Santos, Eds. London, U.K.: Taylor \& Francis, 2022, pp.~337--345.

\bibitem{ferrari2025}
V. Ferrari, A. Maslowska, and T. Gornicz, ``Comparison of regression models for hydrodynamic hull forces,'' in \emph{Technology and Science for the Ships of the Future}, V. Crupi and V. Bucci, Eds. Amsterdam, The Netherlands: IOS Press, 2025, pp.~367--376.

\bibitem{du2021}
T. Du, J. Hughes, S. Wah, W. Matusik, and D. Rus, ``Underwater soft robot modeling and control with differentiable simulation,'' \emph{IEEE Robot. Autom. Lett.}, vol.~6, no.~3, pp.~4994--5001, 2021.

\bibitem{khazoom2024}
C. Khazoom, S. Hong, M. Chignoli, E. Stanger-Jones, and S. Kim, ``Tailoring solution accuracy for fast whole-body model predictive control of legged robots,'' \emph{IEEE Robot. Autom. Lett.}, vol.~9, no.~12, pp.~11074--11081, 2024.

\bibitem{featherstone2008}
R. Featherstone, \emph{Rigid Body Dynamics Algorithms}. New York, NY, USA: Springer, 2008.

\bibitem{carpentier2018}
J. Carpentier and N. Mansard, ``Analytical derivatives of rigid body dynamics algorithms,'' in \emph{Robotics: Science and Systems}, 2018.

\bibitem{carpentier2019}
J. Carpentier, G. Saurel, G. Buondonno, J. Mirabel, F. Lamiraux, O. Stasse, and N. Mansard, ``The Pinocchio C++ library: A fast and flexible implementation of rigid body dynamics algorithms and their analytical derivatives,'' in \emph{Proc. IEEE/SICE Int. Symp. System Integration}, 2019, pp.~614--619.

\bibitem{mastalli2020}
C. Mastalli, R. Budhiraja, W. Merkt, G. Saurel, B. Hammoud, M. Naveau, J. Carpentier, L. Righetti, S. Vijayakumar, and N. Mansard, ``Crocoddyl: An efficient and versatile framework for multi-contact optimal control,'' in \emph{Proc. IEEE Int. Conf. Robotics and Automation}, 2020, pp.~2536--2542.

\bibitem{ramamoorthi2001}
R. Ramamoorthi and P. Hanrahan, ``An efficient representation for irradiance environment maps,'' in \emph{Proc. ACM SIGGRAPH}, 2001, pp.~497--500.

\bibitem{sloan2002}
P.-P. Sloan, J. Kautz, and J. Snyder, ``Precomputed radiance transfer for real-time rendering in dynamic, low-frequency lighting environments,'' in \emph{Proc. ACM SIGGRAPH}, 2002, pp.~527--536.

\bibitem{atkinson2012}
K. Atkinson and W. Han, \emph{Spherical Harmonics and Approximations on the Unit Sphere}. Berlin, Germany: Springer, 2012.

\bibitem{mastalli2022feasibility}
C. Mastalli, W. Merkt, J. Marti-Saumell, H. Ferrolho, J. Sol\`a, N. Mansard, and S. Vijayakumar, ``A feasibility-driven approach to control-limited DDP,'' \emph{Auton. Robots}, vol.~46, no.~8, pp.~985--1005, 2022.

\bibitem{rojas2026odyn}
J. Rojas, A. Papatheodorou, S. Martinez, A. Patrizi, I. Havoutis, and C. Mastalli, ``ODYN: An all-shifted non-interior-point method for quadratic programming in robotics and AI,'' arXiv:2602.16005, 2026.

\bibitem{walker2020}
K. L. Walker, A. A. Stokes, A. Kiprakis, and F. Giorgio-Serchi, ``Impact of thruster dynamics on the feasibility of ROV station keeping in waves,'' in \emph{Proc. OCEANS}, 2020, pp.~1--7.

\bibitem{wendland2005}
H. Wendland, \emph{Scattered Data Approximation}. Cambridge, UK: Cambridge University Press, 2005.

\end{thebibliography}
\end{document}